\documentclass[square, comma]{article}

\PassOptionsToPackage{numbers, compress}{natbib}

\usepackage[preprint]{neurips_2025}

\usepackage[utf8]{inputenc} 
\usepackage[T1]{fontenc}    
\usepackage{hyperref}       
\usepackage{url}            
\usepackage{booktabs}       
\usepackage{amsfonts}       
\usepackage{nicefrac}       
\usepackage{microtype}      
\usepackage{xcolor}         
\usepackage{amsmath} 
\usepackage{multirow}
\usepackage{url}
\usepackage{bbding} 
\usepackage{makecell}
\usepackage{graphicx}
\usepackage{natbib}

\usepackage{algorithm}
\usepackage[noend]{algpseudocode} 
\title{EnerBridge-DPO: Energy-Guided Protein Inverse Folding with Markov Bridges and Direct Preference Optimization}

\author{
  Dingyi Rong \\
  Shanghai Jiao Tong University\\
  \texttt{r892546826@sjtu.edu.cn} \\
  \And
  Haotian Lu \\
  Shanghai Jiao Tong University \\
  \texttt{flick-lu@sjtu.edu.cn} \\
  \AND
  Wenzhuo Zheng \\
  Shanghai Jiao Tong University \\
  \texttt{darkcorvus@sjtu.edu.cn} \\
    \AND
  Fan Zhang \\
  Shanghai Jiao Tong University \\
  \texttt{zf\_521@sjtu.edu.cn} \\
  \And
  Shuangjia Zheng \\
  Shanghai Jiao Tong University \\
  \texttt{shuangjia.zheng@sjtu.edu.cn} \\
  \And
  Ning Liu\\
  Shanghai Jiao Tong University \\
  \texttt{ningliu@sjtu.edu.cn} \\
}

\begin{document}

\maketitle

\begin{abstract}
\label{abstract}
Designing protein sequences with optimal energetic stability is a key challenge in protein inverse folding, as current deep learning methods are primarily trained by maximizing sequence recovery rates, often neglecting the energy of the generated sequences. This work aims to overcome this limitation by developing a model that directly generates low-energy, stable protein sequences. We propose EnerBridge-DPO, a novel inverse folding framework focused on generating low-energy, high-stability protein sequences. Our core innovation lies in: First, integrating Markov Bridges with Direct Preference Optimization (DPO), where energy-based preferences are used to fine-tune the Markov Bridge model. The Markov Bridge initiates optimization from an information-rich prior sequence, providing DPO with a pool of structurally plausible sequence candidates. Second, an explicit energy constraint loss is introduced, which enhances the energy-driven nature of DPO based on prior sequences, enabling the model to effectively learn energy representations from a wealth of prior knowledge and directly predict sequence energy values, thereby capturing quantitative features of the energy landscape. Our evaluations demonstrate that EnerBridge-DPO can design protein complex sequences with lower energy while maintaining sequence recovery rates comparable to state-of-the-art models, and accurately predicts $\Delta \Delta G$ values between various sequences. 
\end{abstract}

\section{Introduction}
\label{introduction}

The inverse protein folding problem—also referred to as inverse folding or fixed-backbone sequence design—seeks to identify amino acid sequences that will reliably fold into a given three-dimensional protein backbone. As one of the central challenges in structural biology and protein engineering, solving this problem not only illuminates the fundamental relationship between sequence and structure, but also underpins the rational design of novel functional proteins and biomolecules \cite{khakzad2023new, chu2024sparks}.

Recent advances in deep learning, coupled with the emergence of large-scale structure-prediction tools and databases, exemplified by AlphaFold 2 \cite{jumper2021highly}, have created unprecedented opportunities for inverse folding. Early methods, such as GraphTrans \cite{wu2021representing}, ProteinMPNN \cite{dauparas2022robust}, and PiFold \cite{gao2022pifold}, which are based on architectures like graph neural networks \cite{scarselli2008graph} and autoregressive models, enforce a one-to-one mapping from structure to sequence, thereby neglecting the inherent diversity of feasible sequences for a given backbone. To explore the one-to-many mapping in the structure-to-sequence space, more recent models leverage advanced iterative strategies \cite{zheng2023structure, gao2023knowledge} or generative models \cite{yi2023graph, zhu2024bridge, wang2024diffusion}. Examples include LM-Design \cite{zheng2023structure}, which utilizes pre-trained language models for sequence optimization; GraDe-IF \cite{yi2023graph}, which employs graph denoising diffusion probabilistic models; and Bridge-IF \cite{zhu2024bridge}, which starts from an information-rich structural prior and utilizes Markov bridges for iterative refinement. Trained on extensive structure–sequence pairings drawn from repositories like the CATH \cite{orengo1997cath} and MPNN datasets \cite{dauparas2022robust}, these models now achieve remarkable performance in both sequence recovery on natural proteins and de novo sequence generation for designed scaffolds.

However, a critical challenge persists: designed sequences must not only be compatible with the target structure but also possess desirable physicochemical properties, particularly low free energy, which is correlated with stability and function \cite{norn2021protein}. Existing mainstream inverse folding models still face several key limitations in generating low-energy sequences: First, current diffusion-based generative models aim to learn a single, often intractable, data distribution. This can lead to a gap between the generated sequences and the true sequence distribution \cite{igashov2023retrobridge, zhu2024bridge, lee2024disco}, making it difficult to efficiently explore the vast sequence space to find candidates that are both structurally consistent and energetically favorable. Second, many advanced inverse folding models are primarily trained by maximizing metrics such as sequence recovery rate or structural similarity. While these models can produce sequences compatible with the target backbone, they generally do not incorporate protein energy stability as a direct optimization objective. Consequently, the generated sequences may not be energetically optimal and could even be unstable. To our knowledge, there is currently a lack of a dedicated inverse folding framework capable of directly and end-to-end optimizing both sequence recovery and low-energy fitness simultaneously, thereby generating sequences intrinsically designed for stability.

\begin{figure}
    \centering
    \includegraphics[width=0.95\linewidth]{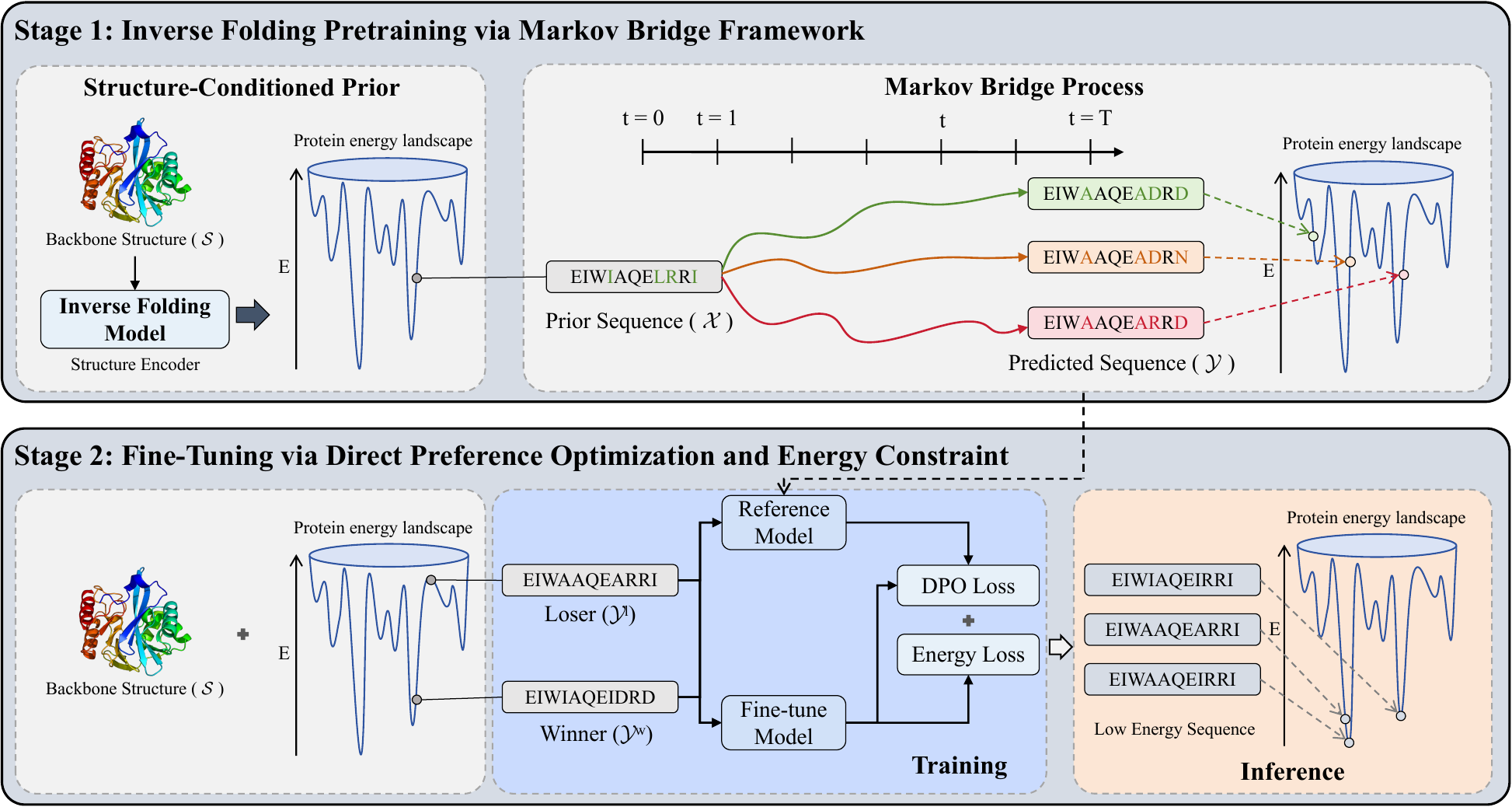}
       \vspace{-0.1in}
    \caption{Overview of EnerBridge-DPO. (a) In Stage 1, we pre-train the model on the structure-to-sequence recovery task using a Markov Bridge framework to enhance its structure-sequence alignment capability and sequence diversity. (b) In Stage 2, we fine-tune the model with Bridge-DPO and energy constraints to guide it toward optimizing within a more energetically stable sequence space.}
    \label{fig:pipeline}
\end{figure}

To address these gaps, we introduce EnerBridge-DPO (Energy-Bridged Direct Preference Optimization), a novel inverse folding framework specifically designed for generating low-energy, high-stability protein sequences. As illustrated in Figure \ref{fig:pipeline}, our method uniquely integrates two key techniques: First, we establish the foundational architecture of the inverse folding model based on a Markov bridge \cite{zhu2024bridge, igashov2023retrobridge} and use DPO \cite{rafailov2023direct} to fine-tune the pre-trained Markov bridge model. Compared to diffusion models, the Markov bridge begins optimization from an information-rich prior sequence related to the target structure, providing DPO with a candidate pool that already possesses structural plausibility. We construct energy-based preference pairs and adapt the DPO training objective to effectively combine with the Markov bridge's generation process. Second, we introduce an explicit energy constraint loss, which directly requires the model to predict the energy values of sequences, enhancing the energy-driven nature of DPO based on prior sequences,. This prompts the model not only to learn the relative energy advantages of sequences but also to understand and fit the quantitative features of the energy landscape. This dual optimization strategy ensures that while EnerBridge-DPO generates low-energy sequences, its internal representations also more closely align with true biophysical principles. Empirical studies show that EnerBridge-DPO outperforms existing baselines on multiple standard benchmarks and excels in designing sequences with low energies.

To summarise, the main contributions of this work are as follows:
\begin{itemize}
\item[$\bullet$] We introduce EnerBridge-DPO, the first inverse folding model that directly utilizes energy constraints within a generative framework to design low-energy protein sequences conforming to specified structural conditions. 
\item[$\bullet$] We utilize energy as a preference to innovatively adapt DPO for effective fine-tuning of the Markov Bridge process. Furthermore, an explicit energy constraint loss is introduced, compelling the model to learn and predict quantitative energy features.
\item[$\bullet$] Experimental results demonstrate that EnerBridge-DPO significantly designs lower-energy and more stable protein complex sequences compared to existing methods, while maintaining comparable sequence recovery and structural validity. The model also accurately predicts $\Delta \Delta G$, highlighting its refined understanding of biophysical principles.
\end{itemize}

\section{Related Work}
\label{related_work}

\subsection{Inverse Protein Folding}
Inverse protein folding aims to identify amino acid sequences that fold into a given three-dimensional protein structure. Early deep learning approaches often employed graph neural networks (GNNs) \cite{ingraham2019generative, gao2022alphadesign, dauparas2022robust, gao2022pifold, tan2022generative, chou2024structgnn}, transformers \cite{ingraham2019generative, hsu2022learning, wu2021representing}, or autoregressive models \cite{ingraham2019generative, hsu2022learning} to learn the mapping from structure to sequence. However, many of these methods model a one-to-one mapping, struggling to capture the inherent sequence diversity for a single backbone (the one-to-many problem) and often facing issues like error accumulation in autoregressive generation. To address these limitations, iterative refinement strategies have emerged. Some methods leverage the knowledge encoded in pre-trained Protein Language Models (PLMs) to refine initially generated sequences, such as LM-Design \cite{zheng2023structure} and KW-Design \cite{gao2023knowledge}. Others employ iterative sampling techniques based on physical models, like ChromaDesign \cite{ingraham2023illuminating} and CarbonDesign \cite{ren2024accurate}, which use methods like MCMC sampling.   

Recently, diffusion models have shown promise in generative tasks. Discrete diffusion models \cite{austin2021structured} have been adapted for sequence generation. GraDe-IF \cite{yi2023graph} pioneered the use of denoising diffusion for inverse folding, conditioning the denoising process on structural information. While diffusion models offer a principled way for iterative refinement and capturing diversity, standard formulations often start from a non-informative prior (e.g., uniform noise), potentially limiting efficiency and the ability to leverage strong structural information directly. Markov Bridge models, such as Bridge-IF \cite{zhu2024bridge}, offer an alternative generative framework. They learn a stochastic process between two distributions, allowing the use of an informative, structure-derived prior sequence as the starting point and progressively refining it towards the target native sequence distribution, potentially offering advantages in sample quality and inference efficiency.

\subsection{Incorporating Physical Constraints and Energy Functions}
A key challenge in protein design is ensuring the physical and chemical viability of generated sequences. Many approaches incorporate physics-based information \cite{norn2021protein, omar2023protein, malbranke2023machine}, often as a post-processing step. For example, generated sequences might be filtered or rescored using energy functions like Rosetta \cite{rohl2004protein} or evaluated using molecular dynamics simulations. While helpful, these two-stage approaches mean the generative model itself isn't directly optimizing for physical properties like energy stability during generation, potentially limiting the effectiveness of the design process. Few methods have successfully integrated energy functions directly into the end-to-end training and optimization loop of deep generative models for inverse folding. Our work, EnerBridge-DPO, aims to bridge this gap by directly optimizing the generative bridge model for lower energy using preference optimization.   

\subsection{Preference Optimization in Generative Models}
Aligning generative models with specific preferences or criteria has been highly successful, particularly in Large Language Models (LLMs). Reinforcement Learning from Human Feedback (RLHF) \cite{christiano2017deep, ziegler2019fine, ouyang2022training, dai2023safe, lee2023rlaif} has become a standard technique, involving training a reward model on human preferences and then optimizing the LLM using RL. More recently, Direct Preference Optimization (DPO) \cite{rafailov2023direct} emerged as a simpler, yet effective alternative that bypasses the need for an explicit reward model, directly optimizing the policy using preference pairs.   

Applying preference alignment to diffusion models is less explored. Some methods fine-tune models on curated, high-quality datasets  or use reward models to guide generation or training. RL-based approaches like DPOK \cite{fan2023dpok} and DDPO \cite{black2023training} have been explored but often face stability issues or limitations in open-vocabulary settings. Diffusion-DPO \cite{wallace2024diffusion} successfully adapted the DPO framework for diffusion models using human preference data, deriving a differentiable objective based on the model's likelihood. 


\section{Methodology}
\label{method}
We present the proposed method in this section. We first give the problem statement in Section \ref{preliminary}. Then we elaborate on the EnerBridge-DPO method, including Markov bridge pre-training in section \ref{bridge}, direct preference optimization (DPO) in section \ref{dpo} and energy constraint in section \ref{energy}.

 \subsection{Preliminary}
\label{preliminary}
The Inverse Folding problem seeks to generate an amino-acid sequence $\mathcal{Y}=(y_1, y_2, ..., y_L)$ that will reliably fold into a given protein backbone structure $\mathcal{S}=(s_1, s_2, ..., s_L)$, where each residue’s coordinates $s_{i} \in \mathbb{R}^{4 \times 3}$ typically include the N, C-$\alpha$, and C atoms (with O atoms optional), and $L$ denotes the length of the backbone. In an ideal design, the proposed sequence should not only reproduce the target backbone under structure prediction but also possess low physical energy and favorable biochemical properties. To formalize this, we treat our generative model as a prior $p_\theta(\mathcal{Y}\mid \mathcal{S})$, and introduce an energy function $\mathcal{E}(\mathcal{S},\mathcal{Y})$ which serves as a likelihood score or potential energy reflecting the physical plausibility of the sequence–structure pair. Under this formulation, the posterior distribution over sequences given the backbone is 

\begin{equation}
    p(\mathcal{Y}\mid \mathcal{S})\propto p_\theta(\mathcal{Y}\mid \mathcal{S})\exp\left(-\beta\mathcal{E}(\mathcal{S},\mathcal{Y})\right)
\end{equation}

where $\beta>0$ governs the relative weight of the energy term. The optimal sequence $\mathcal{Y}^*$—balancing the learned prior and physical constraints—is obtained by maximizing the following objective:
\begin{equation}
    \mathcal{Y}^*=\arg\max_\mathcal{Y}\left[\log p_\theta(\mathcal{Y}\mid \mathcal{S})-\beta\mathcal{E}(\mathcal{S},\mathcal{Y})\right]
\end{equation}
This energy-aware framework underlies modern approaches to inverse folding, enabling the design of sequences that are both structurally faithful and thermodynamically stable.



\subsection{Pre-trained Markov Bridge Model for Inverse Folding}
\label{bridge}
The foundation of EnerBridge-DPO is a generative model designed to capture the complex relationship between protein backbone structures and their corresponding amino acid sequences. This model leverages the Markov Bridge framework  to learn the probabilistic transition from an initial sequence proposal, derived directly from the input structure, to the target native sequence distribution.   
\subsubsection{Structure-Conditioned Prior}
Unlike traditional diffusion models that start from random noise, our approach begins with an informative prior sequence $\mathcal{X}$. This prior is generated by an expressive structure encoder $\mathcal{E}$, which maps the input backbone structure $\mathcal{S}$ to a discrete sequence $\mathcal{X}=\mathcal{E}(\mathcal{S})$. This encoder is pre-trained on large-scale structure-sequence datasets to predict plausible sequences directly from structures. This deterministic mapping provides a strong sequence prior for the subsequent refinement process.   
\subsubsection{Markov Bridge Process}
We establish a discrete-time Markov Bridge process $(\boldsymbol{z}_t)_{t=0}^T$ connecting the distribution of the prior sequence $p_x(\mathcal{X})$ and the distribution of the native sequence $p_y(\mathcal{Y})$. The process starts at $\boldsymbol{z}_0=\mathcal{X}$  and is designed to end at $\boldsymbol{z}_T=\mathcal{Y}$ that satisfies

\begin{equation}
p(\boldsymbol{z}_t|\boldsymbol{z}_0,\boldsymbol{z}_1,\ldots,\boldsymbol{z}_{t-1},\boldsymbol{\mathcal{Y}})=p(\boldsymbol{z}_t|\boldsymbol{z}_{t-1},\boldsymbol{\mathcal{Y}}).
\end{equation}

To pin the process at the end point $\boldsymbol{z}_T=\mathcal{Y}$, we have an additional requirement

\begin{equation}
p(\boldsymbol{z}_T=\mathcal{Y}|\boldsymbol{z}_{T\boldsymbol{-}1},\mathcal{Y})=1.
\end{equation}

We assume that $p(\cdot)$ is categorical distributions with a finite sample space $\{1, . . . , K\}$ and represent data points as one-hot vectors: $\mathcal{X},\mathcal{Y}, \boldsymbol{z}_t \in \{0, 1\}$. The forward process gradually transforms $\mathcal{X}$ towards $\mathcal{Y}$ using a transition matrix $\boldsymbol{Q}_t$:
\begin{equation}
\boldsymbol{Q}_t:=\boldsymbol{Q}_t(\mathcal{Y})=\beta_t\boldsymbol{I}_K+(1-\beta_t)\mathcal{Y}\boldsymbol{1}_K^\top
\end{equation}
where $\beta_t$ is a noise schedule transitioning from $\beta_0=1$ to $\beta_{T\mathrm{-}1}=0$. The core of the generative model lies in learning the reverse process, which approximates the target sequence $\mathcal{Y}$ at each step $t$ given the intermediate state $\boldsymbol{z}_t$ and the structure $\mathcal{S}$.

\subsubsection{Model Architecture and Training}
We utilize a pre-trained Protein Language Model (PLM), such as ESM \cite{rives2021biological}, as the backbone for approximating the reverse bridge process. To effectively condition the PLM on both the time step $t$ and the structural information $\mathcal{S}$, we adapt its architecture using techniques like AdaLN-Bias \cite{peebles2023scalable} and structural adapters (cross-attention). These modifications allow the PLM to leverage its learned evolutionary knowledge while integrating the specific temporal and structural context of the bridge process, all while maintaining parameter efficiency by keeping the base PLM weights frozen. The pre-training minimizes a loss function derived from the Markov Bridge framework, aiming to accurately predict the target sequence $\mathcal{Y}$. We utilize the simplified reparameterized objective function derived in Bridge-IF for more effective training:
\begin{equation}
\mathcal{L}_t(\theta)=\lambda_t\mathbb{E}_{p(\boldsymbol{z}_t|\mathcal{X},\mathcal{Y})}[-v_t\mathcal{Y}^T\log\mathrm{~}\varphi_\theta(\boldsymbol{z}_t,\mathcal{S},t\mathrm{~})]
\end{equation}
where $\mathrm{~}\varphi_\theta(\boldsymbol{z}_t,\mathcal{S},t\mathrm{~})$ is the PLM's prediction of the target sequence, $v_t$ indicates if the token needs refinement, and $\lambda_t$ is a weighting factor. During inference, the model starts with the prior sequence $\boldsymbol{z}_0=\mathcal{E}(\mathcal{S})$ and iteratively refines it using the learned reverse process conditioned on $\mathcal{S}$ and $t$, ultimately generating the final designed sequence $\boldsymbol{z}_T$.

\subsection{Bridge-DPO}
\label{dpo}
We employ Direct Preference Optimization (DPO) \cite{rafailov2023direct} to fine-tune the model, explicitly biasing it towards generating sequences with lower predicted energy states, particularly for protein complexes. This stage aims to integrate physical realism, specifically energy stability, directly into the generative process.

\subsubsection{Preference Data Generation}
Unlike typical DPO applications that rely on human feedback, we leverage the measured energy values as the preference oracle. For a given protein backbone structure $\mathcal{S}$, we generate or select pairs of candidate sequences $(\mathcal{Y}^w, \mathcal{Y}^l)$. A pair is included in our preference dataset $\mathcal{D}_{energy}=\{(\mathcal{S}, \mathcal{Y}^w, \mathcal{Y}^l)\}$ if the experiment measures a lower energy for sequence $\mathcal{Y}^w$ (winning/preferred sequence) compared to sequence $\mathcal{Y}^l$ (losing/less preferred sequence), i.e., $Energy(\mathcal{Y}^w\mid\mathcal{S})<Energy(\mathcal{Y}^l\mid\mathcal{S})$.  

\subsubsection{DPO Objective for Markov Bridges}
We adapt the DPO objective  to the context of our Markov Bridge model. The core idea is to maximize the likelihood of preferred sequences ($\mathcal{Y}^w$) while minimizing the likelihood of less preferred sequences ($\mathcal{Y}^l$), relative to a reference model $\varphi_{ref}$. The reference model $\varphi_{ref}$ is the pre-trained Markov Bridge model obtained from the initial training phase.   
The DPO loss function for fine-tuning our Bridge model $\varphi_{\theta}$ is derived analogously to the objectives in DPO \cite{rafailov2023direct} and Diffusion-DPO \cite{wallace2024diffusion}:   
\begin{equation}
\begin{aligned}\mathcal{L}_\mathrm{Bridge-DPO}(\theta)=&-\mathbb{E}_{(\mathcal{Y}^w,\mathcal{Y}^l)\sim\mathcal{D},t\sim\mathcal{U}(0,T),\boldsymbol{z}_{t}^{w}\sim q(\boldsymbol{z}_{t}^{w}|\mathcal{X}, \mathcal{S}),\boldsymbol{z}_{t}^{l}\sim q(\boldsymbol{z}_{t}^{l}|\mathcal{X}, \mathcal{S})}
\\&\operatorname{log}\sigma (-\beta T\omega(\lambda_{t}) (\|\mathcal{Y}^w-\varphi_\theta(\boldsymbol{z}_t^w,t)\|_2^2-\|\mathcal{Y}^w-\varphi_\mathrm{ref}(\boldsymbol{z}_t^w,t)\|_2^2\big)
\\&-\left(\|\mathcal{Y}^l-\varphi_\theta(\boldsymbol{z}_t^l,t)\|_2^2-\|\mathcal{Y}^l-\varphi_{\text{ref}}(\boldsymbol{z}_t^l,t)\|_2^2\right)\big)\end{aligned}
\end{equation}
Here, $\varphi_\theta(\mathcal{Y}|\mathcal{S})$ represents the probability (or a related measure like likelihood derived from the bridge process) assigned by the model $\varphi_\theta$ to sequence $\mathcal{Y}$ given structure $\mathcal{S}$. $\beta$ is a hyperparameter that controls the strength of the deviation from the reference model $\varphi_{ref}$. A higher $\beta$ imposes a stronger penalty for diverging from the initial pre-trained model, ensuring that the learned preference alignment does not drastically compromise the model's original capabilities. A detailed derivation can be Seen in Appendix \ref{bdpo}.

In the context of Markov Bridge models, the likelihood ratio term can be related to the difference in the model's internal predictions or losses during the bridge process, similar to how Diffusion-DPO relates it to denoising errors. For instance, using the simplified cross-entropy loss formulation, the DPO objective implicitly encourages the model $\varphi_\theta$ to have lower prediction errors (or higher probabilities) for the preferred sequence $\mathcal{Y}^w$ compared to $\mathcal{Y}^l$, relative to the reference model $\varphi_{ref}$, across the bridge timesteps.

\subsection{Energy-constrained Loss}
\label{energy}
To explicitly enforce low‐energy designs, we integrate a BA‑DDG‐based \cite{jiao2024boltzmann} energy constraint into our fine‑tuning stage. First, for each winner-loser pair of a given protein, assuming their backbone structures are identical, we compute the predicted binding free-energy change $\widehat{\Delta\Delta G}$ directly from inverse-folding log-likelihoods via Boltzmann Alignment: 

\begin{equation}
\Delta\Delta G=-k_\mathrm{B}T\cdot\left(\log\frac{p(\mathcal{Y}^\mathrm{winner}_\mathrm{bnd}\mid\mathcal{S}_\mathrm{bnd})}{p(\mathcal{Y}^\mathrm{winner}_\mathrm{unbnd}\mid\mathcal{S}_\mathrm{unbnd})}-\log\frac{p(\mathcal{Y}^\mathrm{loser}_\mathrm{bnd}\mid\mathcal{S}_\mathrm{bnd})}{p(\mathcal{Y}^\mathrm{loser}_\mathrm{unbnd}\mid\mathcal{S}_\mathrm{unbnd})}\right)
\end{equation}

where $k_\mathrm{B}T$ is treated as a learnable scaling factor and 'bnd' means 'bound'. We then define the energy constraint loss as the mean absolute error between predicted and experimental $\Delta\Delta G$ over the labeled set $\mathcal{D}$:
\begin{equation}
\mathcal{L}_\mathrm{energy}(\theta)=\frac{1}{|\mathcal{D}|}\sum_{(s,y_\mathrm{winner},y_\mathrm{loser},\Delta\Delta G)\in\mathcal{D}}\lvert\widehat{\Delta\Delta G}-\Delta\Delta G\rvert
\end{equation}

\subsection{Overall Training Objective}
During the fine-tuning phase of EnerBridge-DPO, the overall training loss $\mathcal{L}_\mathrm{total}(\theta)$ is formulated as a weighted sum of the Bridge-DPO loss ($\mathcal{L}_\mathrm{Bridge-DPO}(\theta)$) and the energy-constrained loss ($\mathcal{L}_\mathrm{energy}(\theta)$):

\begin{equation}
\label{loss}
\mathcal{L}_\mathrm{total}(\theta)=\mathcal{L}_\mathrm{Bridge-DPO}(\theta) + \lambda \mathcal{L}_\mathrm{energy}(\theta)
\end{equation}

Here, the hyperparameter $\lambda$ balances the relative importance of these two objectives.


\section{Experiments}\label{experiments}
In this section, we first give a detailed description of the experimental protocol for this study. We then present comprehensive experiments to demonstrate: 1) The superior performance of our proposed EnerBridge-DPO framework in designing energetically stable protein sequences in inverse folding tasks and its accuracy in protein energy prediction. 2) The effectiveness and contribution of the key individual components within our EnerBridge-DPO framework.
\subsection{Experimental Protocol}

\subsubsection{Datasets} 

We conducted benchmarking on the following datasets:
\begin{itemize}
\item \textbf{MPNN} \cite{dauparas2022robust}: All sequences are clustered at $30 \%$ sequence identity, resulting in 25,361 distinct clusters. We randomly split these clusters into three disjoint sets: a training set (23,358 clusters), a validation set (1,464 clusters), and a test set, ensuring that neither the target chains nor any chains from their biological assemblies appear across multiple splits.

\item \textbf{BindingGym} \cite{lu2024bindinggym}: BindingGym contains 10M mutational data points. For each protein in BindingGYM, we select the top $10 \%$ and bottom $10 \%$ of mutants based on their DMS scores and randomly pair them to construct preference pairs.

\item \textbf{SKEMPI} \cite{jankauskaite2019skempi}: Following the methods of \cite{luo2023rotamer} and \cite{wu2024surface}, we divided the dataset into 3 parts based on structure, ensuring that each part contains unique protein complexes. Based on this division, a three-fold cross-validation process was performed. For each protein complex, we selected mutations from the top $30 \%$ and bottom $30 \%$ ranked by binding energy and randomly paired them to construct preference pairs.
\end{itemize}

\subsubsection{Implementation Details} We pre-trained the model on the MPNN Dataset. Subsequent fine-tuning was performed using preference pairs constructed from the BindingGym Dataset and SKEMPI Dataset. The optimization employed the cosine schedule \cite{nichol2021improved} with number of timestep T = 25. For pre-training, the model was trained for 50 epochs on an NVIDIA 4090 GPU with a batch size of 40,000 residues, an initial learning rate of 0.001, and Adam optimizer \cite{kingma2014adam} with noam learning rate scheduler \cite{vaswani2017attention} was used. During fine-tuning, we maintained the same architecture but reduced the learning rate to $1 \times e^{-5}$, while extending training to 150 epochs for enhanced parameter refinement. Additionally, we set the parameter $\lambda$ in Equation \ref{loss} to 0.5. All experiments are conducted on a computing cluster with CPUs of Intel(R) Xeon(R) Gold 6144 CPU of 3.50GHz and two NVIDIA GeForce RTX 4090 24GB GPU.

\subsubsection{Baselines}
We evaluate the performance of our model on two tasks: protein inverse folding and $\Delta\Delta G$ prediction. For the protein inverse folding task, we compare EnerBridge-DPO against several state-of-the-art baselines, including GraphTrans \cite{wu2021representing}, StructGNN \cite{chou2024structgnn}, GVP \cite{jing2020learning}, GCA \cite{tan2023global}, AlphaDesign \cite{gao2022alphadesign}, ProteinMPNN \cite{dauparas2022robust}, PiFold \cite{gao2022pifold}, LM-Design \cite{zheng2023structure}, GraDe-IF \cite{yi2023graph}, and Bridge-IF \cite{zhu2024bridge}. For the $\Delta\Delta G$ prediction task, we compare our approach with state-of-the-art supervised methods, including DDGPred \cite{shan2022deep}, MIF-Network \cite{yan2020hdock}, RDE-Network \cite{luo2023rotamer}, DiffAffinity \cite{lin2022language}, Prompt-DDG \cite{wu2024learning}, ProMIM \cite{mo2024multi}, Surface-VQMAE \cite{wu2024surface}, and BA-DDG \cite{jiao2024boltzmann}.

\subsubsection{Evaluation}
We use perplexity and recovery rate to evaluate the generation quality of the inverse folding task. Following previous studies \cite{ingraham2019generative}, we report perplexity and median recovery rate under four settings: short proteins (length $<$ 100), medium proteins (100$\leq$length $<$500), long proteins (500$\leq$length $<$1000), and full proteins. Furthermore, we also report the energy of the resulting sequences. In addition to the mean and standard deviation, we also use $\mathrm{ZScore}=e^{\mathrm{mean}_{\mathrm{1}}(x)(\frac{x-\mathrm{mean}_{\mathrm{2}}(x)}{\mathrm{std}_{\mathrm{2}}(x)})}$ to score the models, where $\mathrm{mean}_{\mathrm{1}}$ is the mean value for a specific model across different methods, and $\mathrm{mean}_{\mathrm{2}}$ is the mean value for a specific method across different models.

To comprehensively evaluate the performance of $\Delta\Delta G$ prediction, we use a total of seven overall metrics, including 5 overall metrics: (1) Pearson correlation coefficient, (2) Spearman rank correlation coefficient, (3) minimized RMSE, (4) minimized MAE, (5) AUROC, and 2 per-structure metrics: (6) Per-structure Pearson correlation coefficient and (7) Per-structure Spearman correlation coefficients. 
\subsection{Inverse Folding}
\begin{table}[htbp] 
\centering
\caption{Results comparison on the MPNN dataset. The \textbf{best} and \underline{suboptimal} results are labeled with bold and underlined.}
\label{tab:multichain_results}
\resizebox{\textwidth}{!}{%
  \begin{tabular}{lcccccccc}
\toprule
\multirow{2}{*}{Model} & \multicolumn{4}{c}{Perplexity $\downarrow$} & \multicolumn{4}{c}{Recovery \% $\uparrow$} \\
\cmidrule(lr){2-5} \cmidrule(lr){6-9}
& $L < 100$ & $100 \le L < 500$ & $500 \le L < 1000$ & Full & $L < 100$ & $100 \le L < 500$ & $500 \le L < 1000$ & Full \\

 \midrule
 GraphTrans  & 6.67 & 5.59 & 5.58 & 5.72 & 41.39 & 45.71 & 45.36 & 45.40 \\
 GCA         & 6.41 & 5.33 & 5.30 & 5.45 & 43.45 & 46.93 & 46.52 & 46.58 \\
 SructGNN    & 6.50 & 5.29 & 5.21 & 5.43 & 42.51 & 47.72 & 47.54 & 47.36 \\
 AlphaDesign & 6.77 & 5.33 & 5.21 & 5.49 & 43.62 & 48.04 & 47.81 & 47.81 \\
 GVP         & 6.24 & 4.74 & 4.54 & 4.90 & 45.68 & 51.10 & 51.46 & 50.75 \\
 PiFold      & 6.05 & 4.18 & 3.93 & 4.38 & 48.54 & 55.62 & 56.47 & 55.17 \\
 ProteinMPNN & 5.63 & 4.09 & 3.83 & 4.25 & 50.04 & 57.09 & 59.04 & 57.28 \\
 LMDesign    & 4.60 & 4.06 & 3.99 & 4.06 & 55.07 & 57.18 & 57.99 & 58.14 \\
 Bridge-IF      & \textbf{4.44} & \underline{3.91} & \underline{3.74} & \underline{3.89} & \textbf{58.42} & \underline{60.15} & \underline{60.74} & \underline{60.56} \\
 \midrule
 EnerBridge-DPO & \underline{4.51} & \textbf{3.88} & \textbf{3.68} & \textbf{3.87} & \underline{57.90} & \textbf{60.41} & \textbf{61.14} & \textbf{60.91} \\
\bottomrule
  \end{tabular}%
}
\end{table}

\begin{table}[htbp] 
\centering
\caption{Comparison of energy for sequences designed by different models. The \textbf{best} and \underline{suboptimal} results are labeled with bold and underlined.}
\label{tab:energy_comparison_complexes}
\scalebox{0.9}{
  \begin{tabular}{lccccccc}
\toprule
\multirow{2}{*}{Model} & \multicolumn{2}{c}{FoldX} & \multicolumn{2}{c}{Rosetta}& \multicolumn{2}{c}{BA-Cycle} & \multirow{2}{*}{ZScore $\downarrow$} \\
\cmidrule(lr){2-3} \cmidrule(lr){4-5} \cmidrule(lr){6-7}
& Mean $\downarrow$ & Std $\downarrow$ & Mean $\downarrow$ & Std $\downarrow$ & Mean $\downarrow$ & Std $\downarrow$ \\
\midrule
GraphTrans  & 243.78 & 128.25 & 2813.42 & 3193.46 & 129.19 & 148.83 & 1.82 \\
StructGNN   & 235.33 & 133.79 & 2849.01 & 3212.60 & 121.49 & 153.97 & 1.65 \\
GCA         & 233.97 & 146.12 & 2842.93 & 3206.22 & 126.77 & 148.22 & 1.74 \\
GVP         & 223.32 & 125.85 & 2884.59 & 3283.36 & 125.90 & 141.56 & 1.76 \\
AlphaDesign & 217.01 & 126.02 & 2880.81 & 3273.53 & 128.59 & 151.01 & 1.72 \\
PiFold      & 176.69 & 117.97 & 2814.60 & 3274.96 & 120.65 & 122.20 & 0.86 \\
ProteinMPNN & 163.58 & 109.53 & \textbf{2697.03} & 3362.60 & 112.32 & 125.50 & 0.47 \\
LMDesign    & 202.20 & 295.54 & 3160.13 & 3447.61 & \underline{79.43}  & \textbf{92.83}  & 1.52 \\
Bridge-IF   & \underline{163.02} & \underline{98.63}  & 2807.27 & \textbf{3116.33} & 83.65 & 112.37 & \underline{0.41} \\
\midrule
EnerBridge-DPO & \textbf{130.03} & \textbf{76.31} & \underline{2780.20} & \underline{3157.82} & \textbf{77.46} & \underline{96.79} & \textbf{0.25}  \\
\bottomrule
  \end{tabular}
}
\end{table}

As is shown in Table \ref{tab:multichain_results}, experimental results indicate that EnerBridge-DPO achieves performance comparable to Bridge-IF in terms of sequence recovery rate and perplexity. This suggests that the energy preference fine-tuning via DPO does not negatively impact its fundamental inverse folding fidelity. Despite performing comparably to Bridge-IF on general inverse folding metrics, the core advantage of EnerBridge-DPO lies in the energetic optimization of sequences designed for protein complexes. To validate this, we specially selected 26 protein complex structures with distinct chains from the test set to serve as test cases. For these specific complexes, we employed EnerBridge-DPO and other baseline models to design sequences. Then, using three unsupervised energy prediction models (including FoldX \cite{delgado2019foldx}, Rosetta \cite{alford2017rosetta} and BA-Cycle \cite{jiao2024boltzmann}), we assessed the predicted binding free energy or stability of these designed sequences.

The results, as shown in Table \ref{tab:energy_comparison_complexes}, clearly demonstrate that sequences generated by EnerBridge-DPO for these selected protein complexes have significantly lower predicted energy values than all comparative models. This strongly validates the effectiveness of the energy preference learning introduced during the DPO fine-tuning phase, enabling EnerBridge-DPO to specifically optimize and generate more energetically stable and physicochemically sound protein complex sequences.

\subsection{Protein Energy Prediction}

\begin{table*}[htbp] 
\centering
\caption{Comparison of 3-fold cross-validation on the SKEMPI v2 dataset. The \textbf{best} and \underline{suboptimal} results are labeled with bold and underlined.}
\label{tab:skempi_results_no_supervision_no_newcommand}
\resizebox{\textwidth}{!}{%
\begin{tabular}{lccccccc}
\toprule
\multirow{2}{*}{Method} & \multicolumn{2}{c}{Per-Structure} & \multicolumn{5}{c}{Overall} \\
\cmidrule(lr){2-3} \cmidrule(lr){4-8} 
& Pearson $\uparrow$ & Spear. $\uparrow$ & Pearson $\uparrow$ & Spear. $\uparrow$ & RMSE $\downarrow$ & MAE $\downarrow$ & AUROC $\uparrow$ \\

\midrule
DDGPr\text{ed} & 0.3750 & 0.3407 & 0.6580 & 0.4687 & 1.4998 & 1.0821 & 0.6992 \\
MIF-Network & 0.3965 & 0.3509 & 0.6523 & 0.5134 & 1.5932 & 1.1469 & 0.7329 \\
RDE-Network & 0.4448 & 0.4010 & 0.6447 & 0.5584 & 1.5799 & 1.1123 & 0.7454 \\
DiffAffinity & 0.4220 & 0.3970 & 0.6609 & 0.5560 & 1.5350 & 1.0930 & 0.7440 \\
Prompt-DDG & 0.4768 & 0.4321 & 0.6764 & 0.5936 & 1.5308 & 1.0839 & 0.7567 \\
ProMIM & 0.4640 & 0.4310 & 0.6720 & 0.5730 & 1.5160 & 1.0890 & 0.7600 \\
Surface-VQMAE & 0.4694 & 0.4324 & 0.6482 & 0.5611 & 1.5876 & 1.1271 & 0.7469 \\
BA-DDG & \textbf{0.5603} & \textbf{0.5195} & \underline{0.7319} & \underline{0.6433} & \underline{1.4426} & \textbf{1.0044} & \underline{0.7769} \\
\midrule
EnerBridge-DPO & \underline{0.4981} & \underline{0.4666} & \textbf{0.7487} & \textbf{0.6447} & \textbf{1.4257} & \underline{1.0185} & \textbf{0.7780} \\
\bottomrule
\end{tabular}%
}
\end{table*}
As is shown in Table \ref{tab:skempi_results_no_supervision_no_newcommand}, our results indicate that EnerBridge-DPO achieves $\Delta \Delta G$ prediction performance comparable to that of the BA-DDG predictor. Both methods demonstrate strong correlations with experimental data and low error rates, positioning them at the state-of-the-art for this task. This suggests that through the DPO fine-tuning process, particularly with the inclusion of the $\mathcal{L}_\mathrm{energy}$ term, EnerBridge-DPO has effectively internalized the ability to predict energy changes, rather than merely learning to generate sequences that are likely to be low in energy according to energy-constrained loss. The model's capacity to accurately predict $\Delta \Delta G$ values underscores its refined understanding of the underlying biophysical principles governing protein interactions and stability.

Notably, in the per-structure performance evaluation, while EnerBridge-DPO's performance remains robust, it is slightly outperformed by BA-DDG, a method specifically designed for this task. We hypothesize that this difference may be related to the data sampling strategy employed during our DPO training phase. To construct preference pairs, we sampled from the original data distribution, which might have led to disparities in the amount of data per fold during the three-fold cross-validation training.

\subsection{Ablation Studies}

\begin{table}[htbp]
  \centering
  \caption{Ablation study of key design choices on the MPNN dataset.}
  \label{tab:ablation_studies1}
  \resizebox{\textwidth}{!}{%
  \begin{tabular}{@{}llcccccccc@{}}
    \toprule
     \multirow{2}{*}{w/o DPO} & \multirow{2}{*}{w/o Energy} & \multicolumn{4}{c}{Perplexity $\downarrow$} & \multicolumn{4}{c}{Recovery \% $\uparrow$} \\
    \cmidrule(lr){3-6} \cmidrule(lr){7-10}
     &  &  $L < 100$ & $100 \le L < 500$ & $500 \le L < 1000$ & Full & $L < 100$ & $100 \le L < 500$ & $500 \le L < 1000$ & Full \\
    \midrule
     & \checkmark & 4.59 & 3.99 & 3.97 & 4.06 & 56.05 & 60.10 & 61.10& 60.18 \\
     \checkmark & & 6.10 & 4.04 & 4.08 & 4.16 & 54.81 & 59.50 & 60.12 & 59.52 \\
      & & 4.51 & 3.88 & 3.68 & 3.87 & 57.90 & 60.41 & 61.14 & 60.91 \\
    \bottomrule
    \end{tabular}%
    }
\end{table}

\begin{table*}[htbp] 
\centering
\caption{Ablation study of key design choices on the SKEMPI v2 dataset.}
\label{tab:ablation_studies2}
\resizebox{\textwidth}{!}{%
\begin{tabular}{llccccccc}
\toprule
\multirow{2}{*}{w/o DPO} & \multirow{2}{*}{w/o Energy} & \multicolumn{2}{c}{Per-Structure} & \multicolumn{5}{c}{Overall} \\
\cmidrule(lr){3-4} \cmidrule(lr){5-9} 
& & Pearson $\uparrow$ & Spear. $\uparrow$ & Pearson $\uparrow$ & Spear. $\uparrow$ & RMSE $\downarrow$ & MAE $\downarrow$ & AUROC $\uparrow$ \\

\midrule
& \checkmark & 0.3493 & 0.2975 & 0.4134 & 0.4259 & 1.9005 & 1.3475 & 0.6784 \\
\checkmark & & 0.4893 & 0.4623 & 0.7400 & 0.6227 & 1.4539 & 1.0360 & 0.7720 \\
 &  & 0.4981 & 0.4666 & 0.7487 & 0.6447 & 1.4257 & 1.0185 & 0.7780 \\
\bottomrule
\end{tabular}%
}
\end{table*}
To validate the contributions of the key components within the EnerBridge-DPO framework, we conducted a series of ablation studies. We designed the following model variants for comparison: (1) w/o DPO: This model is fine-tuned using only the energy-constrained loss $\mathcal{L}_\mathrm{energy}$ without the energy preference-based DPO process. (2) w/o Energy: This model is fine-tuned using only the energy preference-based DPO process (via $\mathcal{L}_\mathrm{Bridge-DPO}$) without the explicit energy-constrained loss.


\textbf{w/o DPO}: From Table \ref{tab:ablation_studies1}, we observe that without the DPO process, the core inverse folding metrics are significantly worse. This is because the primary role of DPO is to adjust the generative model's output distribution to align with preferences. With only an energy regression loss, the model might impair the pre-trained model's strong sequence generation capabilities by attempting to forcibly fit energy values. 

\textbf{w/o Energy}: From Table \ref{tab:ablation_studies2}, We observe that without the explicit energy-constrained loss, the model performs very poorly on all $\Delta \Delta G$ prediction metrics. It illustrates that the model might learn to generate lower-energy sequences but will be unable to accurately quantify these energy differences or predict specific $\Delta \Delta G$ values.

\section{Conclusion}
\label{conclusion}
In this paper, we introduce EnerBridge-DPO, a novel framework that addresses the critical challenge of generating energetically stable protein sequences in inverse folding. EnerBridge-DPO uniquely combines Markov bridges with Direct Preference Optimization based on sequence energies, and employs an explicit energy constraint to prompt the model to capture the energy values of generated sequences. Experimental results demonstrate that EnerBridge-DPO successfully designs protein sequences, particularly for complexes, with significantly reduced predicted energies and enhanced stability. Notably, this improved energy performance is achieved while maintaining sequence recovery rates comparable to state-of-the-art methods and accurately predicting $\Delta \Delta G$ values. Future work will explore ensuring the diversity and quality of DPO preference pairs and extending the validation of model performance to a broader range of protein families and more complex structural scenarios.


{
\small

\bibliographystyle{IEEEtran}
\bibliography{reference.bib}

\begin{thebibliography}{10}
\providecommand{\url}[1]{#1}
\csname url@samestyle\endcsname
\providecommand{\newblock}{\relax}
\providecommand{\bibinfo}[2]{#2}
\providecommand{\BIBentrySTDinterwordspacing}{\spaceskip=0pt\relax}
\providecommand{\BIBentryALTinterwordstretchfactor}{4}
\providecommand{\BIBentryALTinterwordspacing}{\spaceskip=\fontdimen2\font plus
\BIBentryALTinterwordstretchfactor\fontdimen3\font minus \fontdimen4\font\relax}
\providecommand{\BIBforeignlanguage}[2]{{%
\expandafter\ifx\csname l@#1\endcsname\relax
\typeout{** WARNING: IEEEtran.bst: No hyphenation pattern has been}%
\typeout{** loaded for the language `#1'. Using the pattern for}%
\typeout{** the default language instead.}%
\else
\language=\csname l@#1\endcsname
\fi
#2}}
\providecommand{\BIBdecl}{\relax}
\BIBdecl

\bibitem{khakzad2023new}
H.~Khakzad, I.~Igashov, A.~Schneuing, C.~Goverde, M.~Bronstein, and B.~Correia, ``A new age in protein design empowered by deep learning,'' \emph{Cell Systems}, vol.~14, no.~11, pp. 925--939, 2023.

\bibitem{chu2024sparks}
A.~E. Chu, T.~Lu, and P.-S. Huang, ``Sparks of function by de novo protein design,'' \emph{Nature biotechnology}, vol.~42, no.~2, pp. 203--215, 2024.

\bibitem{jumper2021highly}
J.~Jumper, R.~Evans, A.~Pritzel, T.~Green, M.~Figurnov, O.~Ronneberger, K.~Tunyasuvunakool, R.~Bates, A.~{\v{Z}}{\'\i}dek, A.~Potapenko \emph{et~al.}, ``Highly accurate protein structure prediction with alphafold,'' \emph{nature}, vol. 596, no. 7873, pp. 583--589, 2021.

\bibitem{wu2021representing}
Z.~Wu, P.~Jain, M.~Wright, A.~Mirhoseini, J.~E. Gonzalez, and I.~Stoica, ``Representing long-range context for graph neural networks with global attention,'' \emph{Advances in neural information processing systems}, vol.~34, pp. 13\,266--13\,279, 2021.

\bibitem{dauparas2022robust}
J.~Dauparas, I.~Anishchenko, N.~Bennett, H.~Bai, R.~J. Ragotte, L.~F. Milles, B.~I. Wicky, A.~Courbet, R.~J. de~Haas, N.~Bethel \emph{et~al.}, ``Robust deep learning--based protein sequence design using proteinmpnn,'' \emph{Science}, vol. 378, no. 6615, pp. 49--56, 2022.

\bibitem{gao2022pifold}
Z.~Gao, C.~Tan, P.~Chac{\'o}n, and S.~Z. Li, ``Pifold: Toward effective and efficient protein inverse folding,'' \emph{arXiv preprint arXiv:2209.12643}, 2022.

\bibitem{scarselli2008graph}
F.~Scarselli, M.~Gori, A.~C. Tsoi, M.~Hagenbuchner, and G.~Monfardini, ``The graph neural network model,'' \emph{IEEE transactions on neural networks}, vol.~20, no.~1, pp. 61--80, 2008.

\bibitem{zheng2023structure}
Z.~Zheng, Y.~Deng, D.~Xue, Y.~Zhou, F.~Ye, and Q.~Gu, ``Structure-informed language models are protein designers,'' in \emph{International conference on machine learning}.\hskip 1em plus 0.5em minus 0.4em\relax PMLR, 2023, pp. 42\,317--42\,338.

\bibitem{gao2023knowledge}
Z.~Gao, C.~Tan, and S.~Z. Li, ``Knowledge-design: Pushing the limit of protein design via knowledge refinement,'' \emph{arXiv preprint arXiv:2305.15151}, 2023.

\bibitem{yi2023graph}
K.~Yi, B.~Zhou, Y.~Shen, P.~Li{\`o}, and Y.~Wang, ``Graph denoising diffusion for inverse protein folding,'' \emph{Advances in Neural Information Processing Systems}, vol.~36, pp. 10\,238--10\,257, 2023.

\bibitem{zhu2024bridge}
Y.~Zhu, J.~Wu, Q.~Li, J.~Yan, M.~Yin, W.~Wu, M.~Li, J.~Ye, Z.~Wang, and J.~Wu, ``Bridge-if: Learning inverse protein folding with markov bridges,'' \emph{arXiv preprint arXiv:2411.02120}, 2024.

\bibitem{wang2024diffusion}
C.~Wang, Y.~Zhou, Z.~Zhai, J.~Shen, and K.~Zhang, ``Diffusion model with representation alignment for protein inverse folding,'' \emph{arXiv preprint arXiv:2412.09380}, 2024.

\bibitem{orengo1997cath}
C.~A. Orengo, A.~D. Michie, S.~Jones, D.~T. Jones, M.~B. Swindells, and J.~M. Thornton, ``Cath--a hierarchic classification of protein domain structures,'' \emph{Structure}, vol.~5, no.~8, pp. 1093--1109, 1997.

\bibitem{norn2021protein}
C.~Norn, B.~I. Wicky, D.~Juergens, S.~Liu, D.~Kim, D.~Tischer, B.~Koepnick, I.~Anishchenko, F.~Players, D.~Baker \emph{et~al.}, ``Protein sequence design by conformational landscape optimization,'' \emph{Proceedings of the National Academy of Sciences}, vol. 118, no.~11, p. e2017228118, 2021.

\bibitem{igashov2023retrobridge}
I.~Igashov, A.~Schneuing, M.~Segler, M.~Bronstein, and B.~Correia, ``Retrobridge: Modeling retrosynthesis with markov bridges,'' \emph{arXiv preprint arXiv:2308.16212}, 2023.

\bibitem{lee2024disco}
D.~Lee, D.~Lee, D.~Bang, and S.~Kim, ``Disco: Diffusion schr{\"o}dinger bridge for molecular conformer optimization,'' in \emph{Proceedings of the AAAI Conference on Artificial Intelligence}, vol.~38, no.~12, 2024, pp. 13\,365--13\,373.

\bibitem{rafailov2023direct}
R.~Rafailov, A.~Sharma, E.~Mitchell, C.~D. Manning, S.~Ermon, and C.~Finn, ``Direct preference optimization: Your language model is secretly a reward model,'' \emph{Advances in Neural Information Processing Systems}, vol.~36, pp. 53\,728--53\,741, 2023.

\bibitem{ingraham2019generative}
J.~Ingraham, V.~Garg, R.~Barzilay, and T.~Jaakkola, ``Generative models for graph-based protein design,'' \emph{Advances in neural information processing systems}, vol.~32, 2019.

\bibitem{gao2022alphadesign}
Z.~Gao, C.~Tan, and S.~Z. Li, ``Alphadesign: A graph protein design method and benchmark on alphafolddb,'' \emph{arXiv preprint arXiv:2202.01079}, 2022.

\bibitem{tan2022generative}
C.~Tan, Z.~Gao, J.~Xia, B.~Hu, and S.~Z. Li, ``Generative de novo protein design with global context,'' \emph{arXiv preprint arXiv:2204.10673}, 2022.

\bibitem{chou2024structgnn}
Y.-T. Chou, W.-T. Chang, J.~G. Jean, K.-H. Chang, Y.-N. Huang, and C.-S. Chen, ``Structgnn: an efficient graph neural network framework for static structural analysis,'' \emph{Computers \& Structures}, vol. 299, p. 107385, 2024.

\bibitem{hsu2022learning}
C.~Hsu, R.~Verkuil, J.~Liu, Z.~Lin, B.~Hie, T.~Sercu, A.~Lerer, and A.~Rives, ``Learning inverse folding from millions of predicted structures,'' in \emph{International conference on machine learning}.\hskip 1em plus 0.5em minus 0.4em\relax PMLR, 2022, pp. 8946--8970.

\bibitem{ingraham2023illuminating}
J.~B. Ingraham, M.~Baranov, Z.~Costello, K.~W. Barber, W.~Wang, A.~Ismail, V.~Frappier, D.~M. Lord, C.~Ng-Thow-Hing, E.~R. Van~Vlack \emph{et~al.}, ``Illuminating protein space with a programmable generative model,'' \emph{Nature}, vol. 623, no. 7989, pp. 1070--1078, 2023.

\bibitem{ren2024accurate}
M.~Ren, C.~Yu, D.~Bu, and H.~Zhang, ``Accurate and robust protein sequence design with carbondesign,'' \emph{Nature Machine Intelligence}, vol.~6, no.~5, pp. 536--547, 2024.

\bibitem{austin2021structured}
J.~Austin, D.~D. Johnson, J.~Ho, D.~Tarlow, and R.~Van Den~Berg, ``Structured denoising diffusion models in discrete state-spaces,'' \emph{Advances in neural information processing systems}, vol.~34, pp. 17\,981--17\,993, 2021.

\bibitem{omar2023protein}
S.~I. Omar, C.~Keasar, A.~J. Ben-Sasson, and E.~Haber, ``Protein design using physics informed neural networks,'' \emph{Biomolecules}, vol.~13, no.~3, p. 457, 2023.

\bibitem{malbranke2023machine}
C.~Malbranke, D.~Bikard, S.~Cocco, R.~Monasson, and J.~Tubiana, ``Machine learning for evolutionary-based and physics-inspired protein design: Current and future synergies,'' \emph{Current Opinion in Structural Biology}, vol.~80, p. 102571, 2023.

\bibitem{rohl2004protein}
C.~A. Rohl, C.~E. Strauss, K.~M. Misura, and D.~Baker, ``Protein structure prediction using rosetta,'' in \emph{Methods in enzymology}.\hskip 1em plus 0.5em minus 0.4em\relax Elsevier, 2004, vol. 383, pp. 66--93.

\bibitem{christiano2017deep}
P.~F. Christiano, J.~Leike, T.~Brown, M.~Martic, S.~Legg, and D.~Amodei, ``Deep reinforcement learning from human preferences,'' \emph{Advances in neural information processing systems}, vol.~30, 2017.

\bibitem{ziegler2019fine}
D.~M. Ziegler, N.~Stiennon, J.~Wu, T.~B. Brown, A.~Radford, D.~Amodei, P.~Christiano, and G.~Irving, ``Fine-tuning language models from human preferences,'' \emph{arXiv preprint arXiv:1909.08593}, 2019.

\bibitem{ouyang2022training}
L.~Ouyang, J.~Wu, X.~Jiang, D.~Almeida, C.~Wainwright, P.~Mishkin, C.~Zhang, S.~Agarwal, K.~Slama, A.~Ray \emph{et~al.}, ``Training language models to follow instructions with human feedback,'' \emph{Advances in neural information processing systems}, vol.~35, pp. 27\,730--27\,744, 2022.

\bibitem{dai2023safe}
J.~Dai, X.~Pan, R.~Sun, J.~Ji, X.~Xu, M.~Liu, Y.~Wang, and Y.~Yang, ``Safe rlhf: Safe reinforcement learning from human feedback,'' \emph{arXiv preprint arXiv:2310.12773}, 2023.

\bibitem{lee2023rlaif}
H.~Lee, S.~Phatale, H.~Mansoor, K.~R. Lu, T.~Mesnard, J.~Ferret, C.~Bishop, E.~Hall, V.~Carbune, and A.~Rastogi, ``Rlaif: Scaling reinforcement learning from human feedback with ai feedback,'' 2023.

\bibitem{fan2023dpok}
Y.~Fan, O.~Watkins, Y.~Du, H.~Liu, M.~Ryu, C.~Boutilier, P.~Abbeel, M.~Ghavamzadeh, K.~Lee, and K.~Lee, ``Dpok: Reinforcement learning for fine-tuning text-to-image diffusion models,'' \emph{Advances in Neural Information Processing Systems}, vol.~36, pp. 79\,858--79\,885, 2023.

\bibitem{black2023training}
K.~Black, M.~Janner, Y.~Du, I.~Kostrikov, and S.~Levine, ``Training diffusion models with reinforcement learning,'' \emph{arXiv preprint arXiv:2305.13301}, 2023.

\bibitem{wallace2024diffusion}
B.~Wallace, M.~Dang, R.~Rafailov, L.~Zhou, A.~Lou, S.~Purushwalkam, S.~Ermon, C.~Xiong, S.~Joty, and N.~Naik, ``Diffusion model alignment using direct preference optimization,'' in \emph{Proceedings of the IEEE/CVF Conference on Computer Vision and Pattern Recognition}, 2024, pp. 8228--8238.

\bibitem{rives2021biological}
A.~Rives, J.~Meier, T.~Sercu, S.~Goyal, Z.~Lin, J.~Liu, D.~Guo, M.~Ott, C.~L. Zitnick, J.~Ma \emph{et~al.}, ``Biological structure and function emerge from scaling unsupervised learning to 250 million protein sequences,'' \emph{Proceedings of the National Academy of Sciences}, vol. 118, no.~15, p. e2016239118, 2021.

\bibitem{peebles2023scalable}
W.~Peebles and S.~Xie, ``Scalable diffusion models with transformers,'' in \emph{Proceedings of the IEEE/CVF international conference on computer vision}, 2023, pp. 4195--4205.

\bibitem{jiao2024boltzmann}
X.~Jiao, W.~Mao, W.~Jin, P.~Yang, H.~Chen, and C.~Shen, ``Boltzmann-aligned inverse folding model as a predictor of mutational effects on protein-protein interactions,'' \emph{arXiv preprint arXiv:2410.09543}, 2024.

\bibitem{lu2024bindinggym}
W.~Lu, J.~Zhang, M.~Gu, and S.~Zheng, ``Bindinggym: A large-scale mutational dataset toward deciphering protein-protein interactions,'' \emph{bioRxiv}, pp. 2024--12, 2024.

\bibitem{jankauskaite2019skempi}
J.~Jankauskait{\.e}, B.~Jim{\'e}nez-Garc{\'\i}a, J.~Dapk{\=u}nas, J.~Fern{\'a}ndez-Recio, and I.~H. Moal, ``Skempi 2.0: an updated benchmark of changes in protein--protein binding energy, kinetics and thermodynamics upon mutation,'' \emph{Bioinformatics}, vol.~35, no.~3, pp. 462--469, 2019.

\bibitem{luo2023rotamer}
S.~Luo, Y.~Su, Z.~Wu, C.~Su, J.~Peng, and J.~Ma, ``Rotamer density estimator is an unsupervised learner of the effect of mutations on protein-protein interaction,'' \emph{bioRxiv}, pp. 2023--02, 2023.

\bibitem{wu2024surface}
F.~Wu and S.~Z. Li, ``Surface-vqmae: Vector-quantized masked auto-encoders on molecular surfaces,'' in \emph{Forty-first International Conference on Machine Learning}, 2024.

\bibitem{nichol2021improved}
A.~Q. Nichol and P.~Dhariwal, ``Improved denoising diffusion probabilistic models,'' in \emph{International conference on machine learning}.\hskip 1em plus 0.5em minus 0.4em\relax PMLR, 2021, pp. 8162--8171.

\bibitem{kingma2014adam}
D.~P. Kingma and J.~Ba, ``Adam: A method for stochastic optimization,'' \emph{arXiv preprint arXiv:1412.6980}, 2014.

\bibitem{vaswani2017attention}
A.~Vaswani, N.~Shazeer, N.~Parmar, J.~Uszkoreit, L.~Jones, A.~N. Gomez, {\L}.~Kaiser, and I.~Polosukhin, ``Attention is all you need,'' \emph{Advances in neural information processing systems}, vol.~30, 2017.

\bibitem{jing2020learning}
B.~Jing, S.~Eismann, P.~Suriana, R.~J. Townshend, and R.~Dror, ``Learning from protein structure with geometric vector perceptrons,'' \emph{arXiv preprint arXiv:2009.01411}, 2020.

\bibitem{tan2023global}
C.~Tan, Z.~Gao, J.~Xia, B.~Hu, and S.~Z. Li, ``Global-context aware generative protein design,'' in \emph{ICASSP 2023-2023 IEEE International Conference on Acoustics, Speech and Signal Processing (ICASSP)}.\hskip 1em plus 0.5em minus 0.4em\relax IEEE, 2023, pp. 1--5.

\bibitem{shan2022deep}
S.~Shan, S.~Luo, Z.~Yang, J.~Hong, Y.~Su, F.~Ding, L.~Fu, C.~Li, P.~Chen, J.~Ma \emph{et~al.}, ``Deep learning guided optimization of human antibody against sars-cov-2 variants with broad neutralization,'' \emph{Proceedings of the National Academy of Sciences}, vol. 119, no.~11, p. e2122954119, 2022.

\bibitem{yan2020hdock}
Y.~Yan, H.~Tao, J.~He, and S.-Y. Huang, ``The hdock server for integrated protein--protein docking,'' \emph{Nature protocols}, vol.~15, no.~5, pp. 1829--1852, 2020.

\bibitem{lin2022language}
Z.~Lin, H.~Akin, R.~Rao, B.~Hie, Z.~Zhu, W.~Lu, A.~dos Santos~Costa, M.~Fazel-Zarandi, T.~Sercu, S.~Candido \emph{et~al.}, ``Language models of protein sequences at the scale of evolution enable accurate structure prediction,'' \emph{BioRxiv}, vol. 2022, p. 500902, 2022.

\bibitem{wu2024learning}
L.~Wu, Y.~Tian, H.~Lin, Y.~Huang, S.~Li, N.~V. Chawla, and S.~Z. Li, ``Learning to predict mutation effects of protein-protein interactions by microenvironment-aware hierarchical prompt learning,'' \emph{arXiv preprint arXiv:2405.10348}, 2024.

\bibitem{mo2024multi}
Y.~Mo, X.~Hong, B.~Gao, Y.~Jia, and Y.~Lan, ``Multi-level interaction modeling for protein mutational effect prediction,'' \emph{arXiv preprint arXiv:2405.17802}, 2024.

\bibitem{delgado2019foldx}
J.~Delgado, L.~G. Radusky, D.~Cianferoni, and L.~Serrano, ``Foldx 5.0: working with rna, small molecules and a new graphical interface,'' \emph{Bioinformatics}, vol.~35, no.~20, pp. 4168--4169, 2019.

\bibitem{alford2017rosetta}
R.~F. Alford, A.~Leaver-Fay, J.~R. Jeliazkov, M.~J. O’Meara, F.~P. DiMaio, H.~Park, M.~V. Shapovalov, P.~D. Renfrew, V.~K. Mulligan, K.~Kappel \emph{et~al.}, ``The rosetta all-atom energy function for macromolecular modeling and design,'' \emph{Journal of chemical theory and computation}, vol.~13, no.~6, pp. 3031--3048, 2017.

\end{thebibliography}
}






\newpage
\appendix
\section*{Appendix}

\section{Algorithms}
\begin{algorithm}
\caption{DPO Fine-tuning for EnerBridge-DPO}
\begin{algorithmic}[1]
\State \textbf{Input:} pre-trained reference model $\varphi_{ref}$, initialize policy model $\varphi_{\theta}$, preference dataset $\mathcal{D}_{energy} = \{(\mathcal{S}, \mathcal{Y}^{w}, \mathcal{Y}^{l})\}$

\State \textbf{Output:} fine-tuned model $\varphi_{\theta}$
\Repeat
    \State Sample $(\mathcal{S}, \mathcal{Y}^{w}, \mathcal{Y}^{l})$ from $\mathcal{D}_{energy}$
    \State Sample timestep $t \sim \mathcal{U}(0,T)$
    \State Generate prior sequence $\mathcal{X} \leftarrow \text{StructureEncoder}(\mathcal{S})$
    \State Generate $z_t^w \sim q(z_t | \mathcal{X}, \mathcal{S}, \mathcal{Y}^w)$ \Comment{Noisy version of winner at $t$}
    \State Generate $z_t^l \sim q(z_t | \mathcal{X}, \mathcal{S}, \mathcal{Y}^l)$ \Comment{Noisy version of loser at $t$}

    \State \Comment{Calculate errors (or negative log-likelihoods from model predictions)}
    \State $\text{err}_{\theta}^w \leftarrow ||\mathcal{Y}^w - \varphi_{\theta}(z_t^w, \mathcal{S}, t)||_2^2$
    \State $\text{err}_{ref}^w \leftarrow ||\mathcal{Y}^w - \varphi_{ref}(z_t^w, \mathcal{S}, t)||_2^2$
    \State $\text{err}_{\theta}^l \leftarrow ||\mathcal{Y}^l - \varphi_{\theta}(z_t^l, \mathcal{S}, t)||_2^2$
    \State $\text{err}_{ref}^l \leftarrow ||\mathcal{Y}^l - \varphi_{ref}(z_t^l, \mathcal{S}, t)||_2^2$

    \State \Comment{DPO loss calculation (Core logic from Eq. 7)}
    \State $\text{diff\_winner} \leftarrow \text{err}_{\theta}^w - \text{err}_{ref}^w$
    \State $\text{diff\_loser} \leftarrow \text{err}_{\theta}^l - \text{err}_{ref}^l$
    \State $\text{dpo\_term} \leftarrow \text{diff\_winner} - \text{diff\_loser}$
    
    \State $L_{DPO} \leftarrow -\log \sigma(-\beta_{dpo} \cdot \text{dpo\_term})$ \Comment{$\sigma$ is the sigmoid function}

    \State Update $\varphi_{\theta}$ using gradient of $L_{DPO}$
\Until{convergence}

\State \Return $\varphi_{\theta}$
\end{algorithmic}
\end{algorithm}

\begin{algorithm}
\caption{Sampling}
\begin{algorithmic}[1]
    \State \textbf{Input:} starting point $\mathcal{S} \sim p_\mathcal{S}$, structure encoder $\mathcal{E}$, neural network $\varphi_{\theta}$
    \State $z_0 \leftarrow \mathcal{E}(\mathcal{S})$
    \For{$t \text{ in } 0, \dots, T-1$}
        \State $\hat{y} \leftarrow \varphi_{\theta}(z_t, t)$ \Comment{Output of $\varphi_{\theta}$ is a vector of probabilities}
        \State $q_{\theta}(z_{t+1}|z_t) \leftarrow \text{Cat}(z_{t+1}; Q_t(\hat{y})z_t)$ \Comment{Approximated transition distribution}
        \State $z_{t+1} \sim q_{\theta}(z_{t+1}|z_t)$
    \EndFor
    \State \textbf{Return} $z_T$
\end{algorithmic}
\end{algorithm}

\section{DPO for Markov Bridge Models}
\label{bdpo}
The goal is to adapt the Direct Preference Optimization (DPO) framework for Markov Bridge models used in protein inverse folding. We have a fixed dataset $\mathcal{D}_{energy} = \{(\mathcal{S}, \mathcal{Y}^w, \mathcal{Y}^l)\}$ where each example contains a protein backbone structure $\mathcal{S}$, a preferred (winner) sequence $\mathcal{Y}^w$, and a dispreferred (loser) sequence $\mathcal{Y}^l$, generated from a reference Markov Bridge model $\phi_{ref}$. We aim to learn a new model $\phi_{\theta}$ that is aligned with these energy-based preferences.

\subsection{Starting Point: The RLHF Objective and Its DPO Reformulation}

The general objective in Reinforcement Learning from Human Feedback (RLHF), which DPO aims to solve more directly, is to optimize a policy $p_{\theta}$ to maximize a reward function $r(c, x_0)$ (where $c$ is context and $x_0$ is generation) while regularizing its deviation from a reference policy $p_{ref}$ using a KL-divergence term:

\begin{equation}
 \max_{p_{\theta}} \mathbb{E}_{c \sim \mathcal{D}_c, x_0 \sim p_{\theta}(x_0|c)} [r(c, x_0)] - \beta \mathbb{D}_{\text{KL}}[p_{\theta}(x_0|c) || p_{ref}(x_0|c)] 
\end{equation}

DPO shows that the optimal solution $p_{\theta}^{*}(x_0|c)$ can be written as:

\begin{equation}
 p_{\theta}^{*}(x_0|c) = \frac{1}{Z(c)} p_{ref}(x_0|c) \exp\left(\frac{1}{\beta} r(c, x_0)\right)
 \end{equation}

This allows rewriting the reward function $r(c, x_0)$ in terms of $p_{\theta}^{*}$ and $p_{ref}$. Substituting this into the Bradley-Terry model for preferences $p(x_0^w > x_0^l | c) = \sigma(r(c, x_0^w) - r(c, x_0^l))$, leads to the DPO loss:
\begin{equation}
 L_{DPO}(\theta) = -\mathbb{E}_{(c, x_0^w, x_0^l)} \left[ \log \sigma \left( \beta \log \frac{p_{\theta}(x_0^w|c)}{p_{ref}(x_0^w|c)} - \beta \log \frac{p_{\theta}(x_0^l|c)}{p_{ref}(x_0^l|c)} \right) \right] 
 \end{equation} 

\subsection{Adapting to Markov Bridge Process}

For generative models like Markov Bridge, the likelihood $p_{\theta}(\mathcal{Y}|\mathcal{S})$ is often intractable as it requires marginalizing over all possible generative paths $z_{0:T}$ (where $z_0 = \mathcal{X}$, the prior sequence from structure $\mathcal{S}$, and $z_T = \mathcal{Y}$, the final generated sequence).

We define the objective over the entire path $z_{0:T}$. The RLHF objective becomes:
\begin{equation}
\max_{p_{\theta}} \mathbb{E}_{\mathcal{S} \sim \mathcal{D}_{\mathcal{S}}, z_{0:T} \sim p_{\theta}(z_{0:T}|\mathcal{S})} [r(\mathcal{S}, \mathcal{Y})] - \beta \mathbb{D}_{\text{KL}}[p_{\theta}(z_{0:T}|\mathcal{S}) || p_{ref}(z_{0:T}|\mathcal{S})]
\end{equation} 

This objective can be optimized directly through the conditional path distribution $p_{\theta}(z_{0:T}|\mathcal{S})$ via a DPO-style loss:

\begin{equation}
\begin{aligned}
L_{\text{DPO-BridgePath}}(\theta) =& -\mathbb{E}_{(\mathcal{S}, \mathcal{Y}^w, \mathcal{Y}^l) \sim \mathcal{D}_{energy}} \left[ \log \sigma \left( \beta \mathbb{E}_{z_{1:T}^w \sim p_{\theta}(z_{1:T}^w|\mathcal{Y}^w, \mathcal{S})} \left[ \log \frac{p_{\theta}(z_{0:T}^w|\mathcal{S})}{p_{ref}(z_{0:T}^w|\mathcal{S})} \right. \right. \right. \\ 
& \left. \left. \left. \qquad -\log \frac{p_{\theta}(z_{0:T}^l|\mathcal{S})}{p_{ref}(z_{0:T}^l|\mathcal{S})} \right] \right) \right] 
\end{aligned}
\end{equation}
where $z_T^w = \mathcal{Y}^w$ and $z_T^l = \mathcal{Y}^l$.

\subsection{Addressing Intractability of Path Sampling and Likelihoods}

Optimizing the above equation is challenging because:
\begin{itemize}
    \item Sampling the full path $z_{1:T} \sim p_{\theta}(z_{1:T}|\mathcal{Y}, \mathcal{S})$ (the reverse bridge process pinned at $\mathcal{Y}$) is inefficient and potentially intractable during training.
    \item The path likelihoods $p_{\theta}(z_{0:T}|\mathcal{S})$ are also intractable.
\end{itemize}

We can substitute the reverse decompositions $p_{\theta}(z_{0:T}|\mathcal{S}) = p(z_0|\mathcal{S}) \prod_{t=1}^T p_{\theta}(z_{t-1}|z_t, \mathcal{S})$ (assuming $z_0$ is fixed given $\mathcal{S}$, so $p(z_0|\mathcal{S})$ might be a delta function or a simple prior $\mathcal{X}$). The log-likelihood ratio for a path then becomes a sum of single-step log-likelihood ratios:
\begin{equation} 
\log \frac{p_{\theta}(z_{0:T}|\mathcal{S})}{p_{ref}(z_{0:T}|\mathcal{S})} = \sum_{t=1}^T \log \frac{p_{\theta}(z_{t-1}|z_t, \mathcal{S})}{p_{ref}(z_{t-1}|z_t, \mathcal{S})}
\end{equation} 

By utilizing Jensen's inequality and assuming uniform sampling of timesteps $t \sim \mathcal{U}(0,T)$, we can get a bound:
\begin{equation}
\begin{aligned}
L_{\text{DPO-BridgeStep}}(\theta) \le& -\mathbb{E}_{(\mathcal{S}, \mathcal{Y}^w, \mathcal{Y}^l), t, z_{t-1,t}^w, z_{t-1,t}^l} \left[ \log \sigma \left( \beta T \left( \log \frac{p_{\theta}(z_{t-1}^w|z_t^w, \mathcal{S})}{p_{ref}(z_{t-1}^w|z_t^w, \mathcal{S})} \right.\right.\right. \\ 
&\qquad \left.\left.\left. - \log \frac{p_{\theta}(z_{t-1}^l|z_t^l, \mathcal{S})}{p_{ref}(z_{t-1}^l|z_t^l, \mathcal{S})} \right) \right) \right] 
\end{aligned}
\end{equation}
Here, $z_{t-1,t}$ are sampled from $p_{\theta}(z_{t-1,t}|\mathcal{Y}, \mathcal{S})$.

\subsection{Approximation using the Forward Process and Model Objective}

Sampling from the reverse joint $p_{\theta}(z_{t-1}, z_t | \mathcal{Y}, \mathcal{S})$ is still difficult. Diffusion-DPO approximates the reverse process $p_{\theta}(x_{1:T}|x_0)$ with the forward noising process $q(x_{1:T}|x_0)$.
For Markov Bridges, we are interested in the model's ability to predict the target sequence $\mathcal{Y}$ (or its properties) from an intermediate state $z_t$. The pre-training objective for the Markov Bridge model $\phi_{\theta}$ in EnerBridge-DPO is given by minimizing a loss related to predicting the target sequence $\mathcal{Y}$ given $z_t, \mathcal{S}, t$:
\begin{equation}
\mathcal{L}_{\text{pretrain}_t}(\theta) = \lambda_{t}\mathbb{E}_{p(z_{t}|\mathcal{X},\mathcal{Y})}[-v_{t}\mathcal{Y}^{T}\log \phi_{\theta}(z_{t},\mathcal{S},t)]
\end{equation}
This is essentially a negative log-likelihood. The key insight from DPO is that the log-likelihood ratio $\log \frac{p_{\theta}(z_{t-1}|z_t, \mathcal{S})}{p_{ref}(z_{t-1}|z_t, \mathcal{S})}$ can be related to the difference in the "energies" or "losses" assigned by the models $p_{\theta}$ and $p_{ref}$. For Bridge-DPO, we adapt this by considering the "error" the model $\phi_{\theta}$ makes in predicting the target sequence $\mathcal{Y}$ from $z_t$. The term $\left\lVert\mathcal{Y} - \phi_{\theta}(z_t, \mathcal{S}, t)\right\rVert_2^2$ represents such an error term (e.g., L2 loss if $\phi_{\theta}$ predicts sequence embeddings, or it can be a proxy for negative log-likelihood if $\phi_{\theta}$ predicts probabilities).

The intermediate states $z_t^w$ and $z_t^l$ are sampled from the forward bridge process, conditioned on the prior sequence $\mathcal{X}$ (derived from $\mathcal{S}$) and implicitly on the target sequences $\mathcal{Y}^w$ and $\mathcal{Y}^l$ respectively. The EnerBridge-DPO simplifies this to $z_t \sim q(z_t|\mathcal{X}, \mathcal{S})$ for the purpose of the DPO loss, which is a common simplification in DPO-like objectives where the "noised" versions are generated from the data points.

\subsection{Final Bridge-DPO Loss Formulation}

Combining these ideas leads to the final Bridge-DPO loss function as:

\begin{equation}
\begin{aligned}\mathcal{L}_\mathrm{Bridge-DPO}(\theta)=&-\mathbb{E}_{(\mathcal{Y}^w,\mathcal{Y}^l)\sim\mathcal{D},t\sim\mathcal{U}(0,T),\boldsymbol{z}_{t}^{w}\sim q(\boldsymbol{z}_{t}^{w}|\mathcal{X}, \mathcal{S}),\boldsymbol{z}_{t}^{l}\sim q(\boldsymbol{z}_{t}^{l}|\mathcal{X}, \mathcal{S})}
\\&\operatorname{log}\sigma (-\beta T\omega(\lambda_{t}) (\|\mathcal{Y}^w-\varphi_\theta(\boldsymbol{z}_t^w,t)\|_2^2-\|\mathcal{Y}^w-\varphi_\mathrm{ref}(\boldsymbol{z}_t^w,t)\|_2^2\big)
\\&-\left(\|\mathcal{Y}^l-\varphi_\theta(\boldsymbol{z}_t^l,t)\|_2^2-\|\mathcal{Y}^l-\varphi_{\text{ref}}(\boldsymbol{z}_t^l,t)\|_2^2\right)\big)\end{aligned}
\end{equation}


This loss encourages $\phi_{\theta}$ to have a relatively lower error for preferred sequences ($\mathcal{Y}^w$) and/or a relatively higher error for dispreferred sequences ($\mathcal{Y}^l$) compared to the reference model $\phi_{ref}$, across the bridge timesteps. This derivation parallels the Diffusion-DPO approach by using model prediction errors as a proxy for the terms in the DPO objective, adapted to the Markov Bridge framework.

\section{Visualization for Protein Folding}
\begin{figure}
    \centering
    \includegraphics[width=0.95\linewidth]{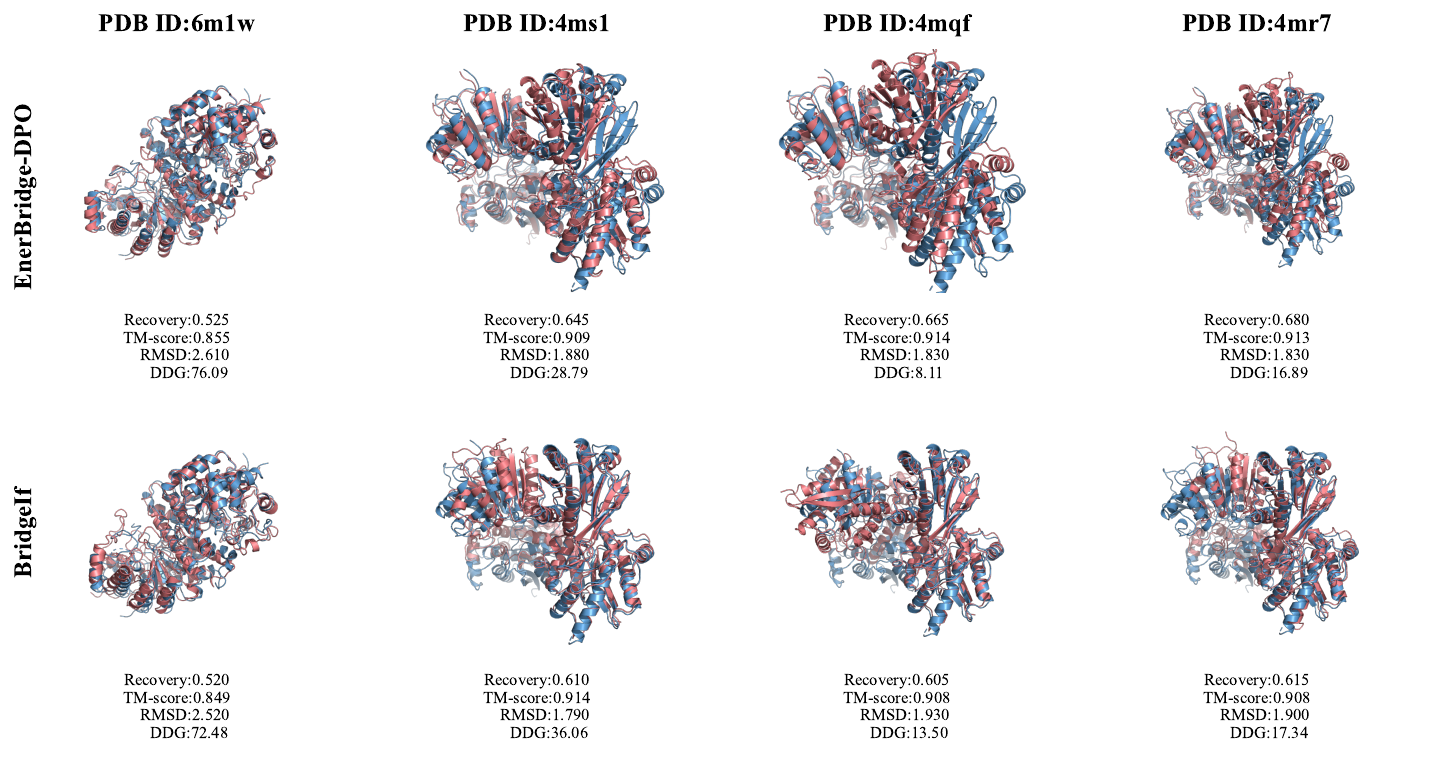}
       \vspace{-0.1in}
    \caption{Folding comparison of designed sequences (in red) and the native sequences (in blue).}
    \label{fig:vis}
\end{figure}
To better show how EnerBridge-DPO designs low-energy sequences, we picked four protein complexes from our test set to look at. Figure \ref{fig:vis} compares the folded structures of sequences designed by EnerBridge-DPO and Bridge-IF with the reference crystal structures. We also used BA-Cycle to predict the energy of these sequences.
\section{Implementation Details}
\subsection{Structure Encoder}
In our framework, the structure-conditioned prior sequence $\mathcal{X}$ is generated by a structure encoder $\mathcal{E}(\mathcal{S})$. In EnerBridge-DPO, we use PiFold as the structure encoder.

\subsection{Preference Data Generation Details}
To construct the preference dataset $\mathcal{D}_{energy}$ for DPO fine-tuning, sequence pairs were carefully selected and constructed from the BindingGym and SKEMPI datasets.
Specifically, for the BindingGym dataset, for each protein we selected the top $10\%$ of mutants with the highest scores as potential 'winners' and the bottom $10\%$ with the lowest scores as potential 'losers'. We then randomly paired sequences from these two groups, resulting in 47,297 million preference pairs. For the SKEMPI v2 dataset, we selected mutant sequences from the top $30\%$ and the bottom $30\%$ and randomly combined them to construct preference pairs, yielding approximately 3,744 preference pairs.
\section{Visualization for $\Delta \Delta G$ Prediction}
\begin{figure}
    \centering
    \includegraphics[width=0.95\linewidth]{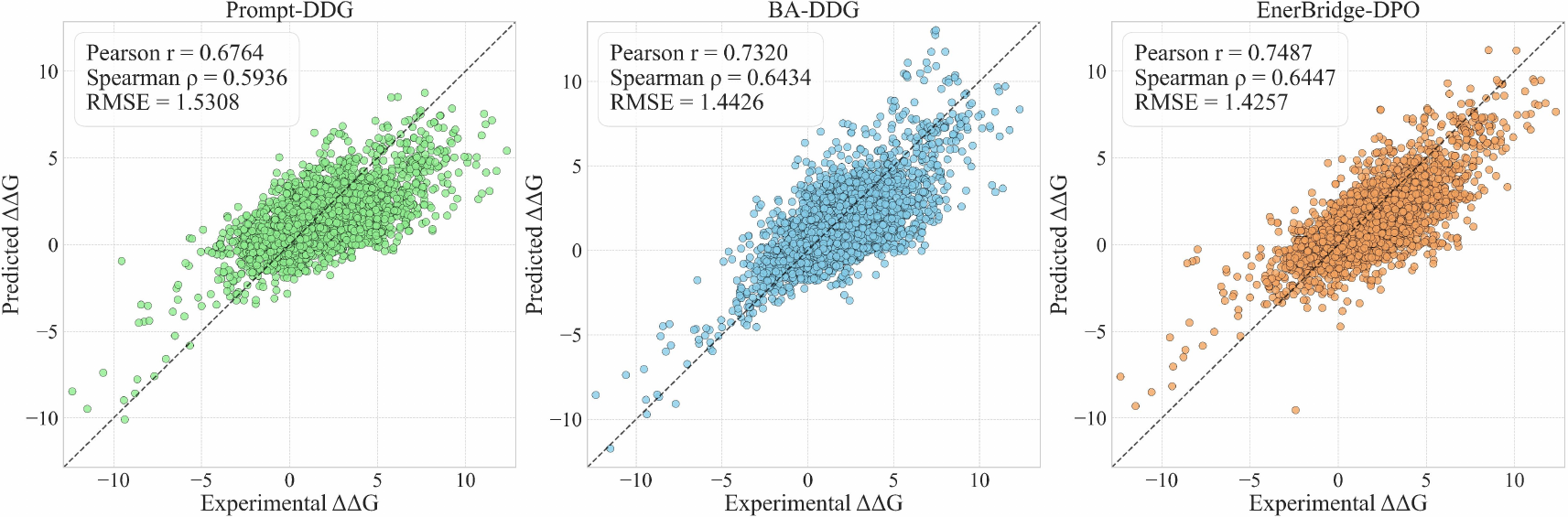}
       \vspace{-0.1in}
    \caption{Comparison of correlations between experimental $\Delta \Delta G$ and predicted $\Delta \Delta G$.}
    \label{fig:sandian}
\end{figure}

Figure \ref{fig:sandian} presents a comparative analysis of the correlation between experimentally determined $\Delta \Delta G$ values and those predicted by different computational methods. The figure comprises three scatter plots, corresponding to Prompt-DDG, BA-DDG, and our proposed EnerBridge-DPO model. In each scatter plot, the x-axis represents the experimental $\Delta \Delta G$ values, while the y-axis denotes the $\Delta \Delta G$ values predicted by the respective model. Ideally, the data points should cluster closely around the diagonal line, indicating a strong agreement between predicted and experimental values.

Visually, the plot for EnerBridge-DPO shows the strongest correlation with experimental data, with its data points appearing more tightly clustered around the diagonal compared to BA-DDG and Prompt-DDG. EnerBridge-DPO achieves higher Pearson and Spearman correlation coefficients and the lowest Root Mean Square Error (RMSE), indicating a smaller overall deviation between its predictions and the experimental results. Collectively, these metrics suggest that EnerBridge-DPO demonstrates superior performance in accurately predicting changes in binding free energy upon mutation compared to the other methods shown.

\section{Broader Impacts}
\label{impacts}
EnerBridge-DPO can significantly accelerate the design of energetically stable proteins, promising advances in developing novel therapeutics and biotechnological solutions, while also deepening our fundamental understanding of protein science. However, thorough experimental validation of all computationally designed proteins is crucial to ensure their real-world safety and efficacy. The responsible development and deployment of this AI-driven technology are paramount to harness its substantial benefits for societal good, particularly in health and sustainability.





\end{document}